# DataParasite Enables Scalable and Repurposable Online Data Curation

Mengyi Sun[1*]
[1]Simons Center for Quantitative Biology, Cold Spring Harbor Laboratory, Cold Spring Harbor, NY, 11724
*To whom correspondence should be addressed: msun@cshl.edu

## Abstract

Many questions in computational social science rely on datasets assembled from heterogeneous online sources, a process that is often labor-intensive, costly, and difficult to reproduce. Recent advances in large language models enable agentic search and structured extraction from the web, but existing systems are frequently opaque, inflexible, or poorly suited to scientific data curation. Here we introduce DataParasite, an open-source, modular pipeline for scalable online data collection. DataParasite decomposes tabular curation tasks into independent, entity-level searches defined through lightweight configuration files and executed through a shared, task-agnostic python script. Crucially, the same pipeline can be repurposed to new tasks, including those without predefined entity lists, using only natural-language instructions. We evaluate the pipeline on multiple canonical tasks in computational social science, including faculty hiring histories, elite death events, and political career trajectories. Across tasks, DataParasite achieves high accuracy while reducing data-collection costs by an order of magnitude relative to manual curation. By lowering the technical and labor barriers to online data assembly, DataParasite provides a practical foundation for scalable, transparent, and reusable data curation in computational social science and beyond.

# Introduction

Data are the bread-and-butter of scientific progress. Across many disciplines, paradigm shifts have followed not from new theories alone, but from the arrival of new forms of data and the ability to analyze them at scale. Examples range from genomics[1] and astronomy[2] to neuroscience[3] and economics[4], where the emergence of large, systematically collected datasets transformed both the questions researchers could ask and the methods they could deploy.

A similar transformation has begun in the social sciences with the rise of computational social science[5]. Enabled by large-scale digital traces, administrative records, and online data, the field has opened new avenues for studying inequality, social structure, collective behavior, and institutional dynamics[6]. Many of the datasets that fuel this work are not released in ready-to-use form, but instead must be assembled from online sources such as organizational websites[7], public records[8], and semi-structured documents[9]. In practice, collecting such data often relies on manual curation or ad hoc scraping pipelines, making the process labor-intensive, expensive, and difficult to scale or reproduce.

Recent advances in large language models (LLMs) offer a potential path toward automating online data collection. Modern LLM-based agents can search the web[10], reason over heterogeneous sources[11], and extract structured information through in-context learning[12]. In particular, emerging forms of agentic search that support multi-turn querying and iterative refinement represent a substantial advance over one-shot retrieval[13]. Despite this promise, there remains a gap between these capabilities and their practical use in computational social science. Existing systems are often closed-source, opaque, or designed as monolithic long-horizon agents that are difficult to adapt, audit, or repurpose for new tasks.

Here we present DataParasite, an open-source, transparent pipeline for scalable and repurposable online data curation. DataParasite is motivated by a simple observation: many datasets of interest in computational social science are naturally tabular, where each row corresponds to an entity (for example, an individual, institution, or event) and each column represents an attribute to be curated. This structure allows online data collection to be decomposed into independent, parallelizable searches at the level of individual entities.

DataParasite implements this idea through a modular design (Fig. 1). Figure 1A shows the core pipeline: a task-specific YAML file defines the schema, prompts, and model settings, while a CSV file specifies the entities to be curated. These inputs are passed to a standalone, task-agnostic execution python script that launches parallel, entity-level curation jobs and compiles the resulting structured outputs into a unified curated database. Figure 1B illustrates how the pipeline can be repurposed via agent-driven configuration. A user initiates a new task using a natural-language request to an off-the-shelf coding agent, which is guided by a lightweight repository playbook. Depending on whether an entity list is provided, the agent either generates the task-specific YAML configuration directly or first identifies relevant online sources and constructs the entity list automatically. In both cases, the resulting configuration files plug directly into the same execution engine shown in Fig. 1A.

We demonstrate the utility of DataParasite on a set of online curation tasks drawn from computational social science. We show that the pipeline can accurately and cost-effectively assemble faculty hiring datasets that are commonly used to study inequality and stratification, while substantially reducing the labor required for data collection. We further demonstrate that the same pipeline can be repurposed, using only natural-language instructions, to distinct tasks with and without predefined entity lists, achieving perfect accuracy in both cases. Together, these results illustrate how DataParasite lowers the barrier to scalable online data collection and provides a practical foundation for data-driven research in computational social science.

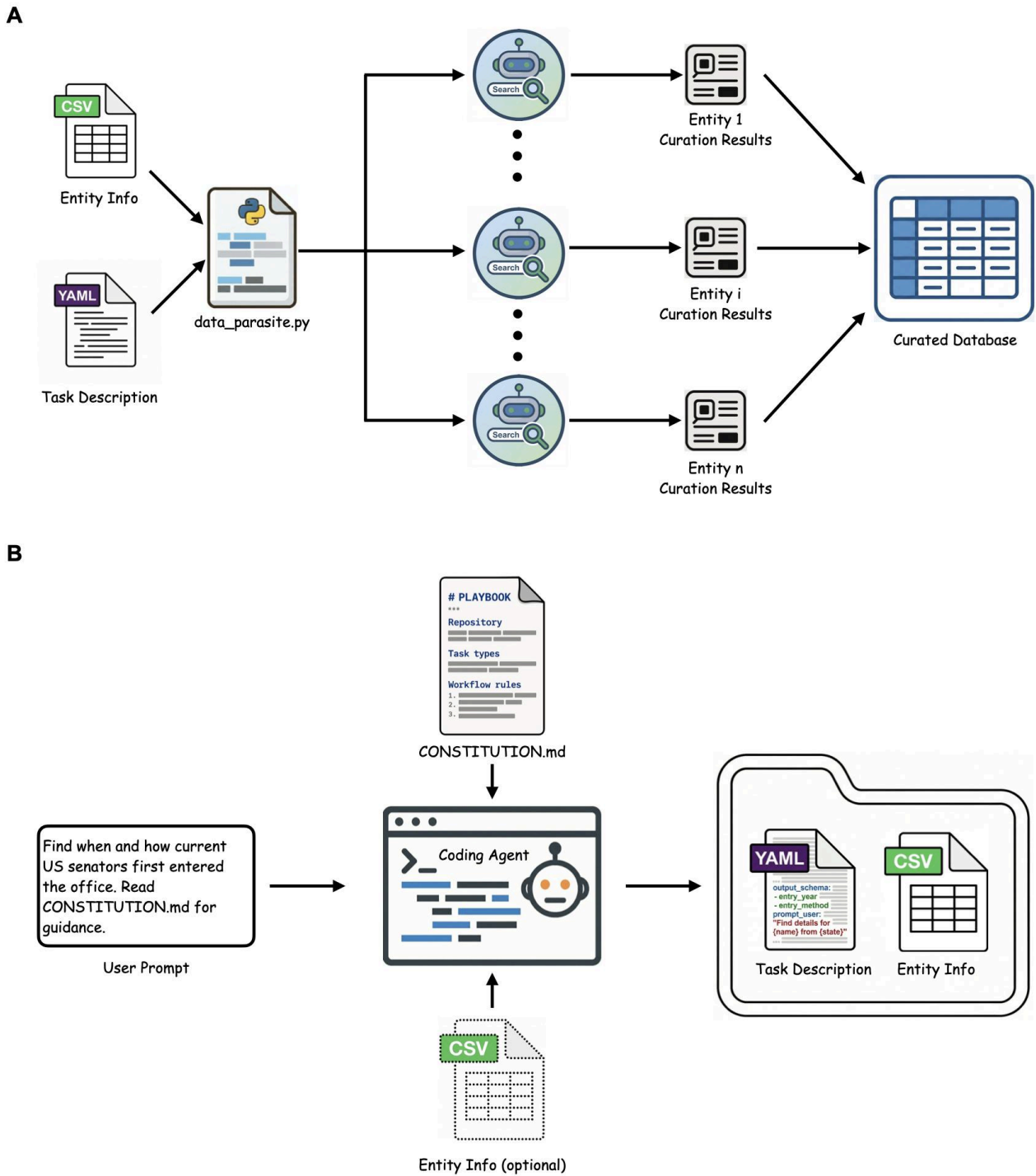

**Fig.1. DataParasite workflow for scalable and repurposable agent-assisted data collection.** **(A)** Core pipeline. A task-specific YAML file ("Task description") defines the schema, prompts, and model settings, and a CSV file ("Entity info") lists the entities to be curated. These inputs are passed to data_parasite.py, which launches parallel LLM-powered curation jobs—one per entity—and compiles the resulting structured outputs into a unified curated database. **(B)** Repurposability via agent-driven configuration. A user can initiate a new task by issuing a natural-language request to a coding agent and directing it to the repository playbook (CONSTITUTION.md). The agent can repurpose the pipeline in **two ways**: *(i)* when a CSV of entities is provided, it generates the corresponding YAML configuration; or *(ii)* when no entity list is provided, it locates the relevant online sources, constructs the entity CSV automatically, and then generates the YAML description. In both cases, the resulting lightweight configuration files plug directly into the same engine shown in (A), enabling rapid reuse of the pipeline across heterogeneous data-collection tasks.

# Results

We evaluated the performance, cost, and flexibility of the DataParasite pipeline on a set of online data-curation tasks that reflect common bottlenecks in computational social science. These tasks require extracting structured information from heterogeneous web sources at scale, where relevant information may be incomplete, inconsistently reported, or difficult to locate automatically.

## Scalable online curation of faculty hiring histories

A central concern across computational social science is inequality in access to opportunity, resources, and status. Faculty hiring has emerged as a canonical empirical lens for studying these dynamics, as patterns of doctoral training and academic placement provide a concrete and well-established instantiation of stratification and mobility in professional labor markets[7]. At the same time, assembling faculty hiring datasets remains a persistent data bottleneck, typically requiring manual curation or customized scraping across heterogeneous departmental and personal websites.

To evaluate whether DataParasite can alleviate this bottleneck, we curated faculty hiring histories for 100 randomly sampled faculty members, extracting three attributes from online sources: doctoral institution, doctoral year, and first tenure-track hiring institution (Fig. 2A). We evaluated accuracy under two regimes: treating "not found" responses as incorrect, and restricting evaluation to entries that were successfully returned as found.

Across all three attributes, DataParasite achieved high accuracy even when "not found" responses were treated as incorrect, and accuracy increased further when analysis was restricted to found entries (Fig. 2A, left). Accuracy was highest for doctoral institution and doctoral year, and remained high for first hire institution. This separation highlights a practical human–AI collaboration regime, in which automated curation resolves the majority of entries with high reliability, while a small residual set of unresolved cases can be delegated to targeted human verification to further improve end-to-end accuracy.

In addition to accuracy, DataParasite substantially reduced marginal data-collection cost. For each faculty record, defined as the cost of curating all three attributes, the distribution of costs was concentrated well below typical human labor rates (Fig. 2A, right). The mean cost per faculty record was approximately $0.09, compared with representative research-assistant costs on the order of $0.66 per record for similar online curation tasks (see **Methods** for estimation details). Beyond cost savings, the pipeline supports parallel processing of many records (within OpenAI rate limits), allowing entire curation tasks to be completed much faster than by a RA. Together, these results demonstrate that DataParasite enables accurate and cost-efficient assembly of faculty placement datasets that have traditionally required substantial manual effort.

## Repurposability across tasks with and without input entity lists

We next evaluated the flexibility of DataParasite by repurposing the same core pipeline to two additional online data-collection tasks using a one-shot natural-language instruction to a coding agent, without modifying the execution engine (Fig. 2B). These tasks were chosen as canonical examples in computational social science and political science where discrete events are frequently used as quasi-exogenous shocks to study long-run outcomes. In both cases, repurposing was achieved by having the coding agent rewrite the task-specific YAML configuration; when an entity list was not provided, the agent additionally located relevant online sources, constructed the corresponding CSV, and then generated the YAML description.

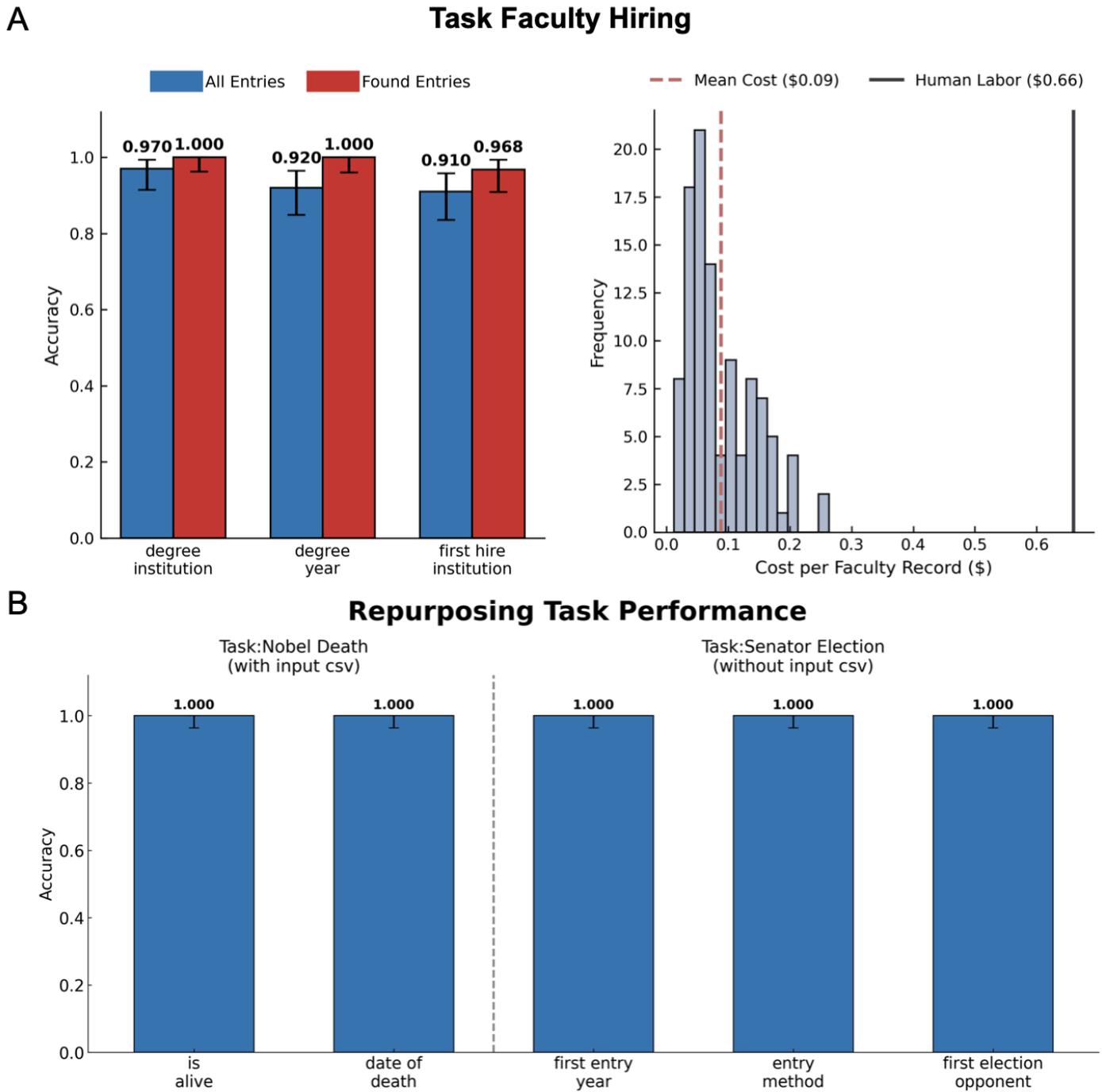

**Fig.2. Accuracy and cost of DataParasite across faculty hiring and repurposed tasks. (A)** Faculty hiring task for 100 faculty. Left panel shows accuracy for online curation of three attributes (degree institution, degree year, first hire institution), comparing evaluation that includes "not found" entries as incorrect (blue) versus evaluation restricted to found entries (red). Bars show point estimates with 95% Clopper–Pearson exact confidence intervals, and numerical values are labeled above bars. Right panel shows the distribution of cost per faculty record, defined as the total cost to curate all three attributes for one faculty member. The dashed line indicates the mean DataParasite cost ($0.09 per record), and the solid line indicates a representative human research assistant cost ($0.66 per record). **(B)** Repurposing task performance (100 entities per task). Left panel (with input csv) shows accuracy for curating whether a Nobel laureate is alive and, if deceased, the date of death. Right panel (without input csv) shows accuracy for curating each current US senator's first entry year, entry method, and first election opponent. Bars show point estimates with 95% Clopper–Pearson exact confidence intervals.

**Task 1 (with input CSV)** focuses on death events of highly influential individuals, which have been widely used to study the causal impact of "great individuals" on peer effect[14], direction of innovation [15], and institutional trajectories[16]. Given an input CSV of 100 randomly sampled Nobel laureates, the coding agent generated a YAML configuration to curate online whether each individual was alive and, if deceased, the date of death.

**Task 2 (without input CSV)** represents a more challenging setting, as the set of relevant entities must first be identified online before any curation can occur. Initial entry into legislative office is a consequential political event not only for individual careers[17] but also for the represented electorate[18], as early electoral outcomes can generate persistent differences in political representation through incumbency advantage[18]. In this task, the coding agent first identified the set of current US senators from online sources, constructed the entity list automatically, and then generated the YAML configuration to curate each senator's first entry year, entry method, and first election opponent.

Despite this increase in task complexity, DataParasite achieved perfect accuracy across all queried attributes in both tasks (Fig. 2B). Because both tasks involve facts that are widely documented online, one potential concern is that high accuracy could arise from model memorization rather than genuine online curation. To rule this out, we conducted a baseline evaluation in which the same model used for curation (gpt-5-mini) was instructed to answer the same questions using its internal knowledge only, with web search explicitly disabled. Under this setting, performance degraded substantially. For the Nobel death task, accuracy for determining whether a laureate was alive was 44%, and for all death-date queries the model returned not_found. For the senator election task, the model likewise returned not_found for all queried attributes. These results indicate that the strong performance of DataParasite arises from active online data collection and verification, rather than from memorized knowledge embedded in the language model.

## Discussion

DataParasite provides a practical tool for scalable online data curation that prioritizes ease of use, repurposability, and low overhead. Across multiple tasks, the pipeline enables accurate and cost-efficient extraction of structured information from heterogeneous web sources, without requiring task-specific code or bespoke scraping pipelines. By expressing data-collection tasks through simple configuration files and executing them through a shared pipeline, DataParasite allows researchers to move quickly from a research question to a working data-collection workflow, and to adapt that workflow to new tasks with minimal effort.

Several limitations suggest directions for future improvement. First, although DataParasite is fully open-source, the current implementation relies on commercial language model APIs, which may limit accessibility and obscure intermediate reasoning trajectories. Supporting open-source models[19,20], including those deployable on edge devices[20–22], would improve transparency and broaden access. Second, web search is currently the dominant cost driver (>90% of the cost), accounting for the majority of total expense. While built-in search tools offer robustness, integrating local retrieval mechanisms such as headless browsing could substantially reduce costs, albeit with practical constraints imposed by rate limits and anti-scraping measures. Third, the pipeline currently treats entities as independent search units, even though relevant information is often clustered across entities. Leveraging such a structure could improve efficiency and accuracy, but requires new approaches to coordinating search and evidence reuse. Finally, while our demonstrations focus on computational social science, similar online curation challenges arise in many scientific domains, including biology[23], medicine[24], and chemistry[25], making broader applications of DataParasite a natural next step.

By making online data curation easier to set up, cheaper to run, and simpler to reuse, DataParasite lowers the barrier to data-driven research and enables researchers to focus on substantive questions rather than data assembly.

## Methods
### Data
For the Faculty Hiring Task, we constructed the input CSV by randomly sampling 100 computer scientists from the United States, China, the United Kingdom, and Germany using the CSRankings dataset (https://github.com/emeryberger/CSrankings/tree/gh-pages). For each individual, the ground-truth Ph.D. institution, graduation year, and first hiring institution were manually curated from online sources and used to evaluate the DataParasite curation results.

For the Nobel Death Task, the list of Nobel laureates and associated prize information was obtained from a Kaggle dataset (https://www.kaggle.com/datasets/imdevskp/nobel-prize). We randomly sampled 100 laureates and manually curated their vital status (alive or deceased), including the date of death when applicable, from online sources. These ground-truth records were then compared against the DataParasite curation results.

For the Senator Election Task, no input list was provided a priori. Instead, a coding agent (we used Cursor-Agent; tests with Codex produced similar results) automatically retrieved the current list of U.S. senators and their corresponding states from online sources and generated the downstream task YAML file. We manually verified the retrieved senator list against the official Senate directory (https://www.senate.gov/general/resources/pdf/senators_phone_list.pdf), and the curated career information was cross-checked against Wikipedia.

### Implementation of DataParasite
DataParasite is implemented using the OpenAI Python SDK. In addition to the core logic described in the main text, the pipeline automatically records telemetry, including token usage, web-search tool invocation counts, and associated costs. The implementation supports switching among different OpenAI models; however, all analyses presented in this study were conducted using gpt-5-mini. The source code for DataParasite is available at: https://github.com/mengysun/DataParasite.

### Cost Estimation of Human Labor
The average cost of human labor is estimated based on a recent study with comparable task complexity, namely determining the gender and Ph.D. year of a given author[26]. That study reports that a research assistant can curate approximately 30 records per hour, which is consistent with our own experience when manually curating ground-truth data online. Assuming a typical U.S. research assistant wage of approximately $20 per hour, this corresponds to an estimated cost of $0.66 ($20/30) per record.

### Prompts for Repurposing Tasks
The prompt for the Nobel Death Task is as follows:
"I have compiled a list of Nobel Laureates in the folder `nobel_winner_death` under the task directory. Your task is to determine whether each individual is still alive—and if not, when they passed away. Before launching the task, make sure to read the `@CONSTITUTION.md` file carefully."

The prompt for the Senator Election Task is as follows:
"Collect data on all current (2025) U.S. senators, including the year they first entered the Senate (appointed or elected) and, if elected, the final opponent they defeated in the first successful election. Name the task `senator_election`. Read the `@CONSTITUTION.md` carefully before launching the task."